\documentclass[10pt,twocolumn,letterpaper]{article}

\usepackage{cvpr}
\usepackage{times}
\usepackage{epsfig}
\usepackage{graphicx}
\usepackage{amsmath}
\usepackage{amssymb}
\usepackage[labelfont={bf,small},font=small]{caption}
\usepackage{subfig}
\usepackage{multirow}
\usepackage{multicol}
\usepackage{iac_pkg}
\usepackage{balance}

\usepackage[pagebackref=true,breaklinks=true,letterpaper=true,colorlinks,bookmarks=false]{hyperref}
\usepackage[capitalise]{cleveref}
\cvprfinalcopy % *** Uncomment this line for the final submission

\def\cvprPaperID{17} % *** Enter the CVPR Paper ID here

\ifcvprfinal\fi
\begin{document}

%%%%%%%%% TITLE
%\title{On the Importance of Temporal Information \\ for Face Manipulation Detection in Videos}
\title{Recurrent Convolutional Strategies\\ for Face Manipulation Detection in Videos}
%\title{Recurrent-Convolution Approach to DeepFake Detection -- State-Of-Art Results on FaceForensics++}

\author{Ekraam Sabir, Jiaxin Cheng, Ayush Jaiswal, Wael AbdAlmageed, Iacopo Masi, Prem Natarajan \\
USC Information Sciences Institute, Marina del Rey, CA, USA\\
{\tt\small \{esabir, chengjia, ajaiswal, wamageed, iacopo, pnataraj\}@isi.edu}
% For a paper whose authors are all at the same institution,
% omit the following lines up until the closing ``}''.
% Additional authors and addresses can be added with ``\and'',
% just like the second author.
% To save space, use either the email address or home page, not both
}

\maketitle
\thispagestyle{empty}

%%%%%%%%% ABSTRACT
\begin{abstract}
The spread of misinformation through synthetically generated yet realistic images and videos has become a significant problem, calling for robust manipulation detection methods. 
Despite the predominant effort of detecting face manipulation in still images, less attention has been paid to the identification of tampered faces in videos by taking advantage of the temporal information present in the stream. Recurrent convolutional models are a class of deep learning models which have proven effective at exploiting the temporal information from image streams across domains.
We thereby distill the best strategy for combining variations in  these models along with domain specific face preprocessing techniques through extensive experimentation to obtain state-of-the-art performance on publicly available video-based facial manipulation benchmarks.
Specifically, we attempt to detect Deepfake, Face2Face and FaceSwap tampered faces in video streams. Evaluation is performed on the recently introduced FaceForensics++ dataset, improving the previous state-of-the-art by up to 4.55\% in accuracy.
\end{abstract}
\section{Introduction}\label{sec:intro}
A spate of recent incidents have increased the scrutiny of online misinformation \cite{ferrara_disinformation_2017,allcott_social_2017}. This has spurred research on both analysis and detection of misinformation \cite{vosoughi_spread_2018,ruchansky_csi:_2017}. Misinformation can be manifested in different ways --- direct manipulation of information or presentation of unmanipulated content in a misleading context. Digital image manipulation such as copy-move and splicing \cite{wu_busternet:_2018,wu_deep_2017} are examples of deliberate manipulation, while image repurposing \cite{jaiswal_multimedia_2017,sabir_deep_2018,jaiswal_aird:_2019} is an example of misleading context. Out of the wide range of manipulations on different modalities, \emph{automatic} manipulation of digital content has recently gained attention. 
In particular, facial manipulations lately became very popular as a way for disseminating false information or even to libel celebrities or very well-known people~\cite{deepfake_defame}. It is not surprising that faces are preferred over other objects, since the face is used everyday in our society as a means for human communication and associating information with identity.

Prior to the rise of deep learning, a malicious attacker would \emph{manually} prepare counterfeit media either using Adobe Photoshop \footnote{\href{https://www.adobe.com/products/photoshopfamily.html}{www.adobe.com/products/photoshopfamily.html}} or GIMP\footnote{\href{https://www.gimp.org/}{www.gimp.org}}, making the process tedious, but nowadays the scenario has changed drastically: before machine learning came in the process, tools only aided the user in content creation, whilst currently, machine-learning--aided tools create content without manual intervention. The user is just required to ``babysit'' the training process, most of the time using easy-to-use Graphical User Interfaces (GUI)~\cite{DeepFaceLab,deepfakes_non_2019}.

More precisely, fueled by the recent success of Generative Adversarial Networks (GANs)~\cite{goodfellow2014generative} along with the availability of graphical processors (GPUs), any amateur user is capable of producing completely synthetic yet hyper-realistic content. The realism achieved by the aforementioned machine learning components is so high that even humans have difficulties in assessing whenever a face picture is naturally captured with a camera or, artificially produced. This statement is even more true for \emph{single still images}.
For instance, websites such as \href{https://thispersondoesnotexist.com/}{thispersondoesnotexist.com} offer evidences of the level of realism reached by the aforementioned tools. The level of artefacts produced by such systems is so subtle that the only clues to assess if a face is real or fake are (i) subtle inconsistency in the hairs --- too straight, with disconnected strands or simply unnatural, (ii) unnaturally asymmetric face, (iii) weird teeth, and more importantly and most of the time, (iv) other more clear inconsistencies are not localized on the face yet in the background. These latter artefacts are peculiar of generative networks that render the entire head along with the background e.g., StarGAN~\cite{choi2018stargan}.

While we believe that it is important to develop detectors to pick up those salient irregularities in still images~\cite{zhou2017two,afchar_mesonet:_2018} to stem the spread of false information, additional, orthogonal important features for detecting manipulations in videos are captured by the temporal coherence inherently present in a video stream.
In this sense, face manipulation in videos has recently gained interest upon still images. The increased proliferation of fake videos may be attributed to two reasons: 
\begin{enumerate}
    \item Transposing a person's identity or expression with somebody else's, is easier now given the off-the-self availability of machine-learning-aided face swapping or face reenactment tools~\cite{thies_face2face:_2016,deepfakes_non_2019,nirkin2018face,DeepFaceLab}.
    \item A video is more likely to be believable rather than a still image since it demonstrates the activity in progress and also, in principle, it requires a much painstaking effort in manipulating \emph{coherently} all the frames.
\end{enumerate}

In light of above observations and considering that face manipulation generation tools~\cite{thies_face2face:_2016,deepfakes_non_2019,nirkin2018face,DeepFaceLab} do not enforce temporal coherence in the synthesis process and perform manipulations on a frame-by-frame basis, we propose to leverage temporal artefacts as a means for indication of abnormal faces in a video stream. Unlike state-of-the-art fake detection methods~\cite{zhou2017two,afchar_mesonet:_2018}, we explore recurrent convolutional models to exploit temporal discrepancies for improving upon the current practice. To remove other confounding factors related to the rigid motion of the face in a video, we also explore face alignment methods. We optimize a deep learning model architecture over these two factors which gives state-of-the-art performance in face manipulation detection accuracy.

The rest of the manuscript is organized as follows: \cref{sec:related} presents the related work on temporal processing with deep models, recent face manipulation benchmarks along with manipulation detection techniques recently published. \cref{sec:method} explains the proposed methodology comprising of face preprocessing, feature extraction and finally prediction based on a bi-direction recurrent model. The experimental evaluation is presented in~\cref{sec:expts} with quantitative results on FaceForensiscs++ (FF++)~\cite{rossler_faceforensics++:_2019}. \cref{sec:conclusions} concludes the paper.
\section{Related Work}\label{sec:related}

\begin{figure*}[!t]
\centering
\fbox{
	\includegraphics[width=\linewidth]{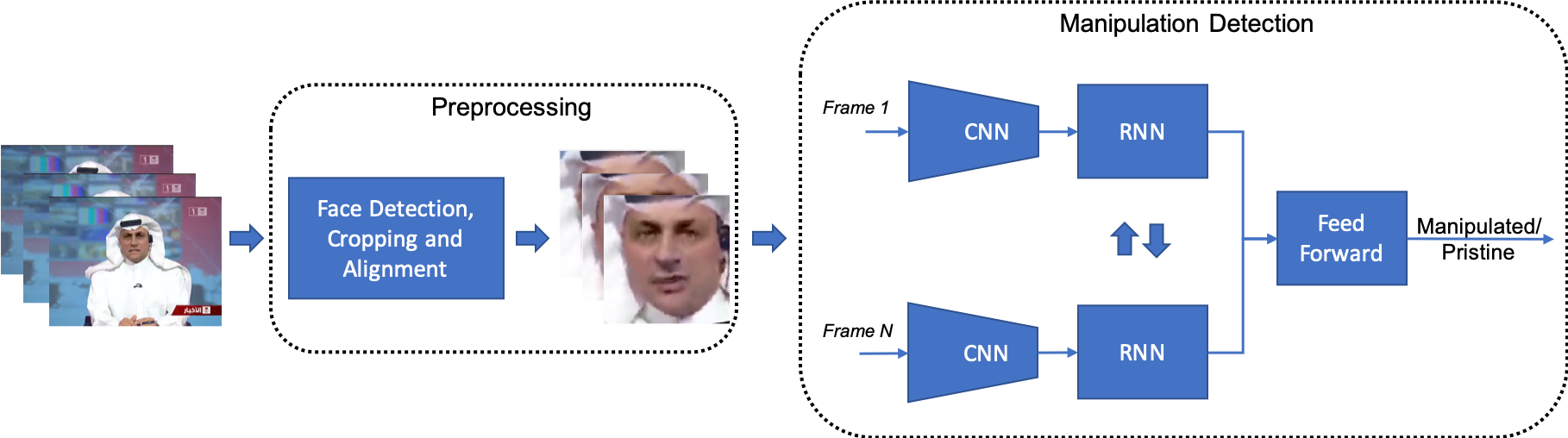}
	}
\caption{The overall pipeline is a two step process. The first step detects, crops and aligns faces on a sequence of frames. The second step is manipulation detection with our recurrent convolutional model.}
\label{Fig:Model}
\end{figure*}

\minisection{Video processing with deep models} Activity recognition in videos has well developed literature and can be used to gain insights into processing videos, taking advance of temporal information. There are three major approaches in this area. The first line of research develops from a two stream network~\cite{simonyan_two-stream_2014}, where an RGB video frame and its optical flow version are processed in two separate branches in the network followed by a fusion mechanism. This former strategy of video processing aims to capture temporal information and motion across frames with the usage of optic flow. The second is a single stream network supported by recurrent convolutional layers: perceptual knowledge of the content in each frame is gathered using a separate convolutional neural network (CNN) that extracts high-level semantic features, while a recurrent model is trained on top of those features to perform decisions over the temporal dimension~\cite{donahue_long-term_2015}. The third line of development resides in 3D convolutions~\cite{tran2015learning,ji20133d} as a local building block in the network to learn rich spatio-temporal features.
With respect to all the previous mentioned strategies, we argue that methods based on two-stream architecture are effective in action recognition but not relevant to capture the subtle flickering artefacts that a generator may produce in a video; on the other hand, 3D convolutions could be better suited for this purpose but they greatly increase the number of learned filters. For all these reasons, we use a main backbone CNN encoder and capture temporal anomaly in face appearance using a recurrent model. Recurrent models have been widely used in computer vision for face landmark detection~\cite{lai2018deep}, face age progression~\cite{wang2016recurrent} and even face parsing~\cite{liu2017face}; nevertheless, to the best of our knowledge they have not been employed before for video manipulation, with the only exception of~\cite{guera2018deepfake}.

\minisection{Face manipulation benchmarks} Regrettably, compelling datasets for face manipulation detection and evaluation in videos had been lacking in the community; some previous attempts~\cite{zhou2017two} generated face swapped images using an iOS app and a open-source face swap software using still images; though Zhou \etal used this set for evaluation, it did not provide manipulation for video streams. 

Video-based face manipulation became available for the community with the recent release of FaceForensics~\cite{rossler_faceforensics:_2018}, followed by its extended and improved version called FaceForensics++~\cite{rossler_faceforensics++:_2019}. FaceForensics (FF) released Face2Face manipulated videos~\cite{thies_face2face:_2016}. FaceForensics++ (FF++) is an extension of FF, further augmenting the collection with \textit{Deepfake}~\cite{deepfakes_non_2019} and \textit{FaceSwap}~\cite{marek_3d_2019} manipulations. The set comprises of 1,000 videos organized into a single split where 720 videos are reserved for training and 140 videos used for validation and the same amount for test. All videos are collected from Youtube and capture ``talking heads'' or anchor men/women presenting news. In general, the subject faces the camera and cooperates with it.

Regarding the manipulation types, \textit{Deepfakes} are generated via a two-branch autoencoder network. Deepfake system is a identity-swapping method and trained given two subjects. In this base version, a single autoencoder needs to be trained for a pair of individuals. The autoencoder shares a single encoder for compressing the information while two decoders reconstruct each one of the subject's images. At test time, the two decoders are inverted to obtain the final face identity manipulation also known as face swapping; on the other hand, \textit{FaceSwap} is a graphics-based approach to attain the same objective of swapping the identity of subjects. Unlike \textit{Deepfakes}, \textit{FaceSwap} can be applied to an unlimited number of subjects though is more prone to severe artefacts if some of its inside modules fail. Finally, \textit{Face2Face}~\cite{thies_face2face:_2016} is a graphics-based approach yet achieves very realistic facial reenactment. Reenactment transfers the expression and pose of a source character into a target video, while the identity of the target subject remains unvaried: i.e. it offers the same functionality of animating a pre-recorded video.  For more details, we refer to~\cite{rossler_faceforensics++:_2019}. %There are three versions of this dataset based on compression: high quality (no compression), low quality (mild compression) and very low quality (heavy compression). %% maybe move this into the explanation of the dataset.

\minisection{Face Manipulation Detection} Copy-move and splicing detection datasets and methods are abundant for images~\cite{pandey_passive_2014,wu2018image, wu_busternet:_2018,mantranet2019} and less dominant in videos~\cite{jia2018coarse}. However, due to the recent emergence of face manipulation problem the literature is relatively sparse. Rossler \etal~\cite{rossler_faceforensics++:_2019} introduced baselines implemented through existing methods and thus trained an XceptionNet~\cite{chollet_xception:_2017} architecture to establish state-of-the-art on the FF+ benchmark. In the meantime, MesoNet~\cite{afchar_mesonet:_2018} introduces two CNN based architectures for face manipulation detection: they take a mesoscopic approach to manipulation detection, which combines information from both low level (microscopic) and high level (macroscopic) features. Their main novelty resides in the MesoInception block that extends the Inception module~\cite{szegedy2016rethinking} with the usage of dilated convolution~\cite{yu2017dilated}.
Zhou \etal trained a GoogLeNet based architecture for detecting tampered face in still images, in synergy with another model working with steganalysis features and trained with triplet loss. The two models' scores are combined through late fusion.
In order to improve transfer learning of features Cozzolino \etal~\cite{cozzolino2018forensictransfer} leverage the abundant data present for face manipulation by using FF++ collections and adapted the model to newer domain using \emph{ForensicTransfer}: an encoder-decoder architecture provided with an activation loss measures the activation of the latent space vector distinctively for the real class and the fake class. The closest prior work to ours is~\cite{guera2018deepfake}: unlike them, we experiment with face alignment as a means for removing confounding factors in detecting facial manipulations and we make use of bidirectional recurrency rather than just mono-directional. Additionally, we attempt to detect multiple types of face manipulation while the previous work is centered around deepfake detection.

\section{Method}\label{sec:method}

\begin{table*}[t]
    \centering
    \begin{tabular}{|c|c|c|c|c|c|c|c|c|}
        \hline
         \multirow{3}{*}{Manipulation} & \multirow{3}{*}{Frames} & \multirow{2}{*}{FF++} & \multirow{3}{*}{ResNet50} & \multirow{3}{*}{DenseNet} & \multirow{2}{*}{ResNet50} & \multirow{2}{*}{DenseNet} & ResNet50 & DenseNet \\
         & & \multirow{2}{*}{\cite{rossler_faceforensics++:_2019}} & & & \multirow{2}{*}{+ Alignment} & \multirow{2}{*}{+ Alignment} & + Alignment & + Alignment\\
         & & & & & & & + BiDir & + BiDir \\
         \hline
         \hline
         \multirow{2}{*}{Deepfake} & 1 & 93.46 & 94.8 & 94.5 & 96.1 & 96.4 & - & - \\
         \cline{2-9}
         & 5 & - & 94.6 & 94.7 & 96.0 & 96.7 & 94.9 & \textbf{96.9} \\
         \hline
         \hline
         \multirow{2}{*}{Face2Face} & 1 & 89.8 & 90.25 & 90.65 & 89.31 & 87.18 & - & - \\
         \cline{2-9}
         & 5 & - & 90.25 & 89.8 & 92.4 & 93.21 & 93.05 & \textbf{94.35} \\
         \hline
         \hline
         \multirow{2}{*}{FaceSwap} & 1 & 92.72 & 91.34 & 91.04 & 93.85 & 96.1 & - & - \\
         \cline{2-9}
         & 5 & - & 90.95 & 93.11 & 95.07 & 95.8 & 95.4 & \textbf{96.3} \\
         \hline
    \end{tabular}
    \caption{Accuracy for manipulation detection across all manipulation types. DenseNet with alignment and bidirectional recurrent network is found to perform best. FF++ \cite{rossler_faceforensics++:_2019} is the baseline in these experiments.}
    \label{tab:Table1}
\end{table*}

\begin{table}[t]
    \centering
    \begin{tabular}{|c|c|c|c|}
    \hline
        \multirow{3}{*}{Manipulation} & \multirow{3}{*}{Base} & \multicolumn{2}{c|}{Variation} \\
        \cline{3-4}
        & & Spatial & Multi \\
        & & Transformer & Recurrence \\
        \hline
        \hline
         Deepfake & \textbf{96.9} & 91.7 & 94.4 \\
         \hline
         Face2Face & \textbf{94.35} & 87.46 & 89.9 \\
         \hline
         FaceSwap & \textbf{96.3} & 93.2 & 94.8 \\
         \hline
    \end{tabular}
    \caption{Results on using variations to the recurrent convolutional architecture. Both spatial-transformer networks and multi-recurrent networks exhibit a decline in performance.}
    \label{tab:Table2}
\end{table}

The overall approach for manipulation detection is a two step process: cropping and alignment of faces from video frames, followed by manipulation detection over the preprocessed facial region. We explain both the steps in detail in this section.

\subsection{Face preprocessing}\label{sec:preproc}
For cropping the face region, we use the masks provided by~\cite{rossler_faceforensics++:_2019}, generated using computer graphics~\cite{thies_face2face:_2016}. Since it has been shown in~\cite{masi_learning_2019,masi2019face} that face alignment is beneficial for face recognition, we employ face alignment here as well.
In particular, we have experimented with two techniques for alignment: (i) explicit alignment using facial landmarks, where a reference coordinate system and the tightness of the face crop are decided \textit{a priori} and all the faces are aligned to this reference coordinate system so that any \emph{rigid} motion of the face is compensated, and (ii) implicit alignment that uses a Spatial Transformer Network (STN)~\cite{jaderberg2015spatial,wu2017recursive} based on an affine transformation. In the latter case, the network predicts the alignment parameters, conditioned on the input image, thus may learn to ``zoom'' on particular parts of the face, as necessary to minimize the expected loss in the training set. 

In both cases the core idea is that we want the recurrent convolutional model to take as input a face ``tubelet''~\cite{kalogeiton2017action,kang2017object}, which is a sequence of spatio-temporal, tightly aligned face crops across video frames.

\minisection{Landmark-based alignment} Face images are aligned using a simple similarity transformation (four degrees of freedom), compensating for isotropic scale, in-plane rotation, and 2D translation. Most of the faces in the dataset are near-frontal and thus, it was sufficient to employ an accurate yet fast landmark detector method \cite{kazemi_one_2014} implemented through \cite{king_dlib-ml:_2009}.
Though \cite{kazemi_one_2014} returns dense landmarks on the face, we select only a set of seven sparse points located on the most discriminative features of the faces (corners of the eyes, the tip of the nose, and corners of the mouth). Following the similarity transformation, faces are aligned with a loose crop at a $224\times 224$ resolution.

\minisection{Spatial Transformer Network} A spatial transformer network (STN) \cite{jaderberg2015spatial} performs spatial alignment of data with learnable affine transformation parameters. It can be inserted between feature maps in deep learning networks. It comprises three components: a localization net, a grid generator and a sampler. The localization net predicts the affine transformation parameters and the grid generator and sampler warp the input feature map with the affine parameters to produce the output feature maps. They have been shown to focus on spatially important areas of the input feature map for improved performance \cite{jaderberg2015spatial}.

\begin{figure*}[!t]
    \centering
    \subfloat[]{\includegraphics[width=0.45\linewidth]{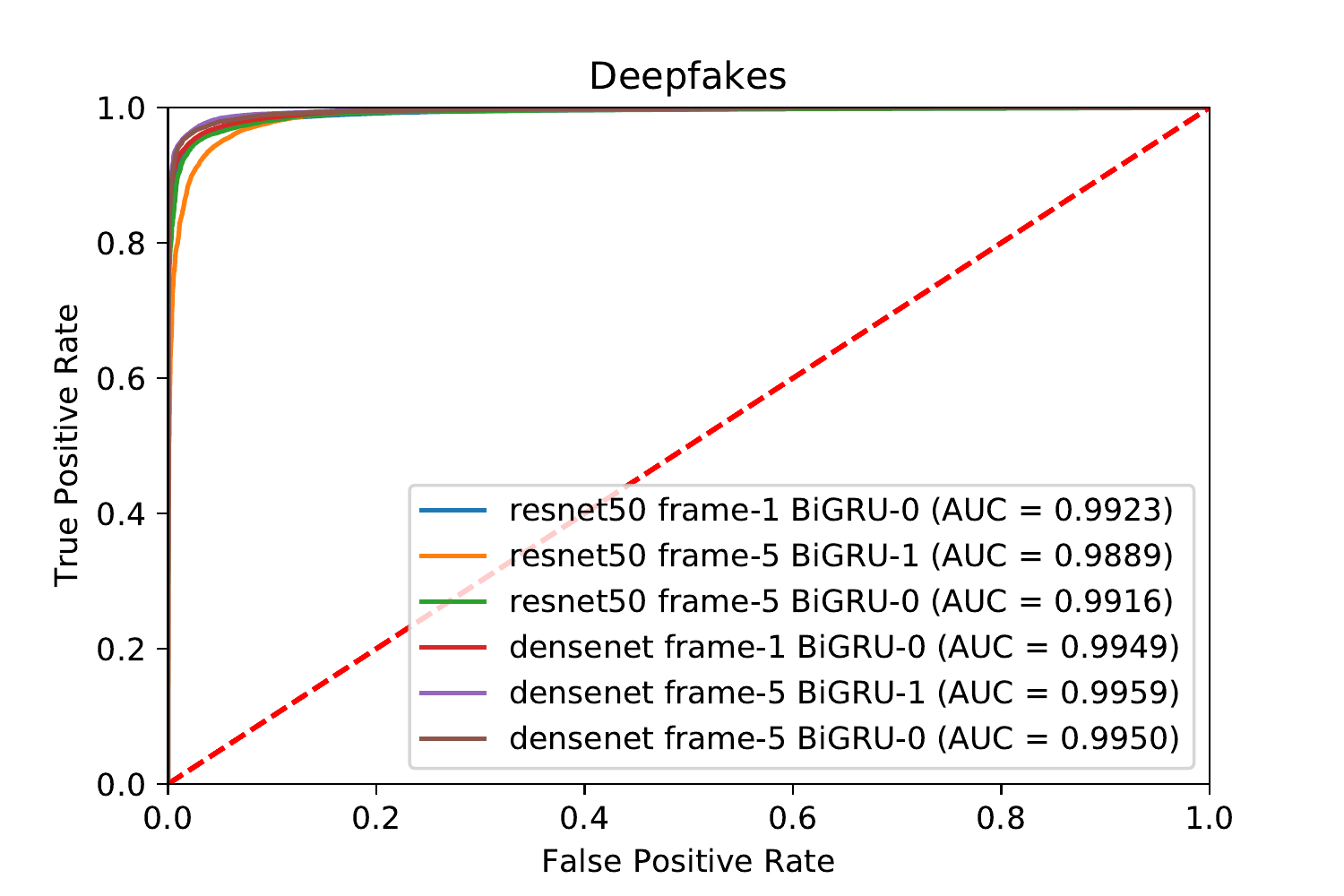} } 
    \subfloat[]{\includegraphics[width=0.45\linewidth]{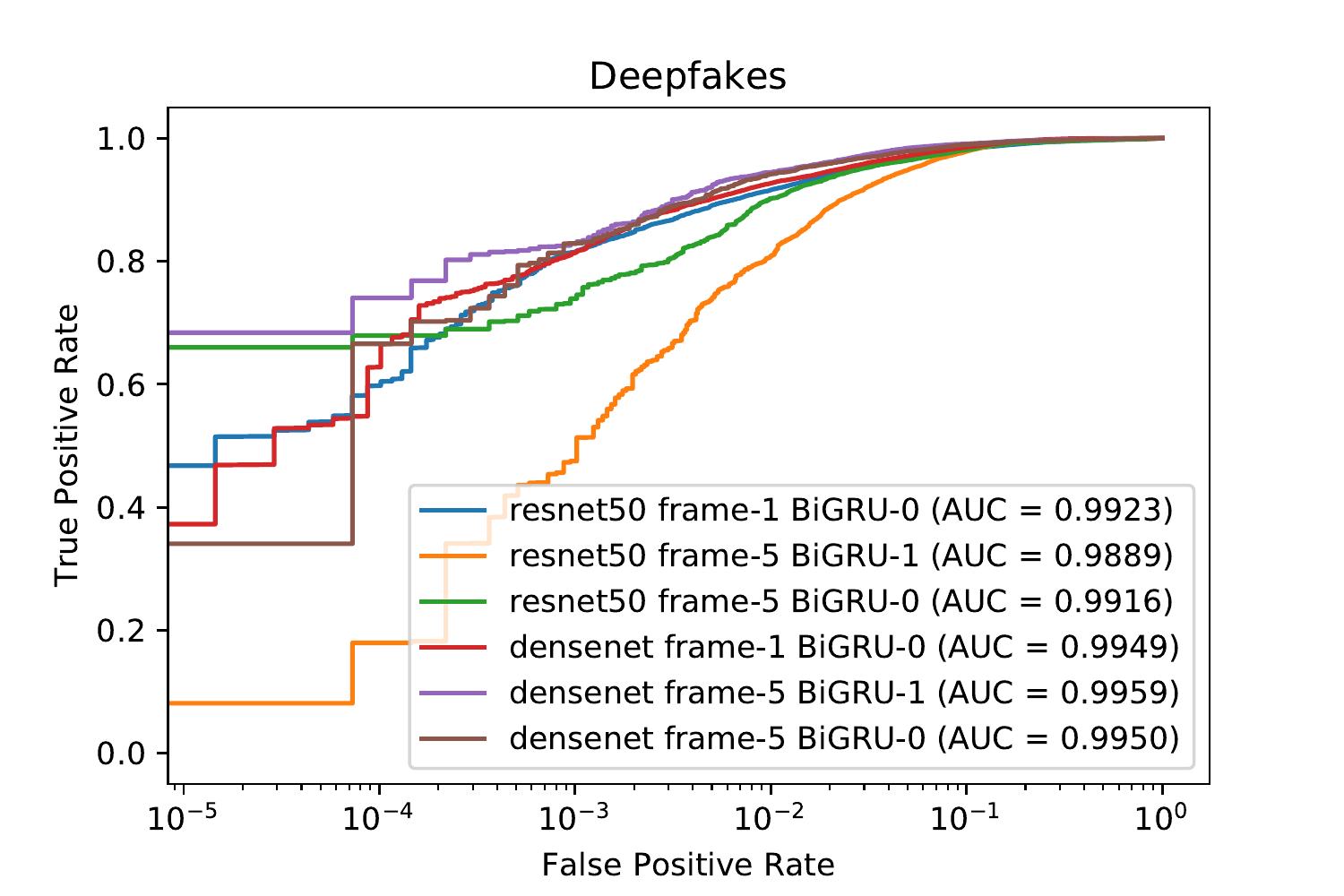}}\\
    \subfloat[]{\includegraphics[width=0.45\linewidth]{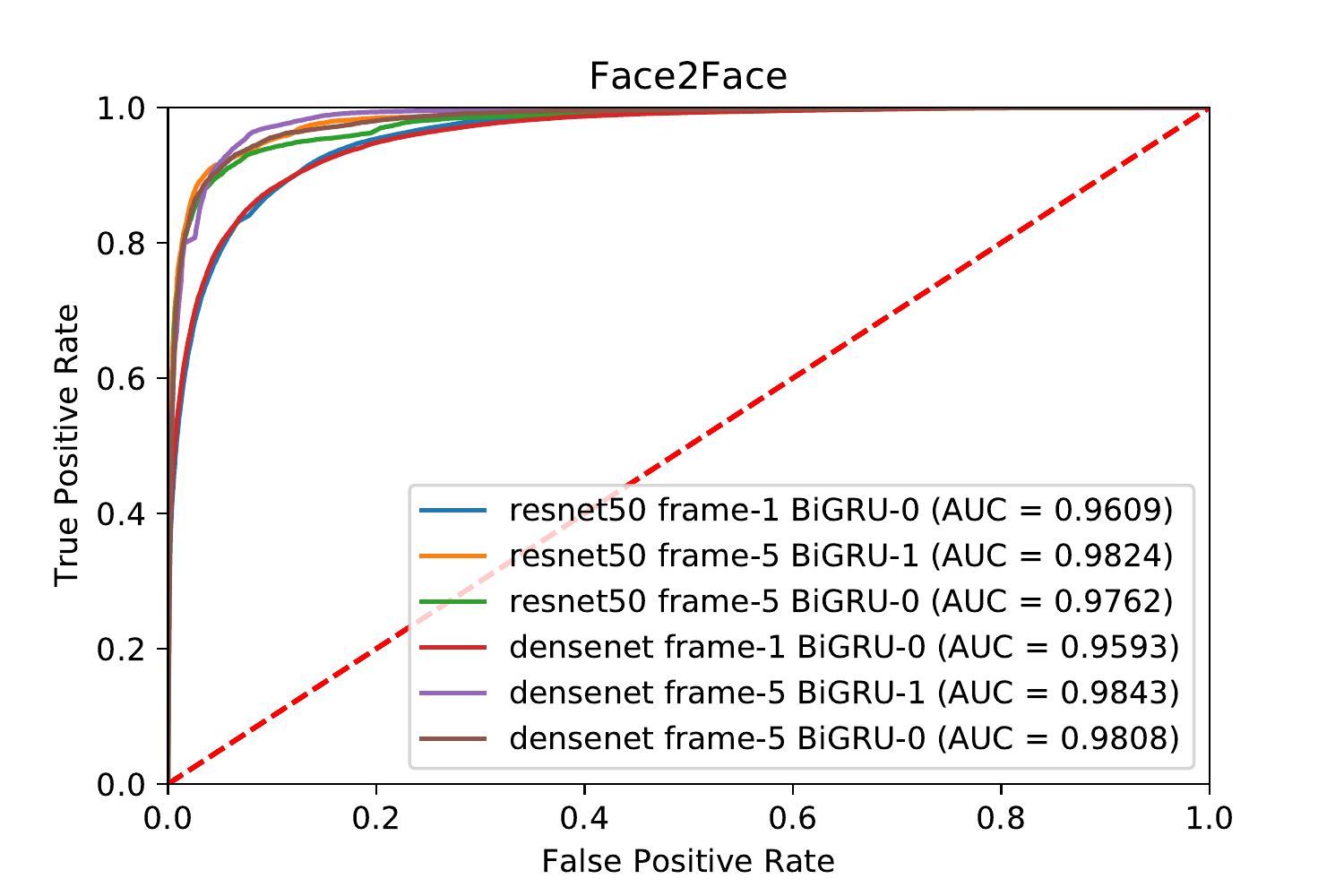} } 
    \subfloat[]{\includegraphics[width=0.45\linewidth]{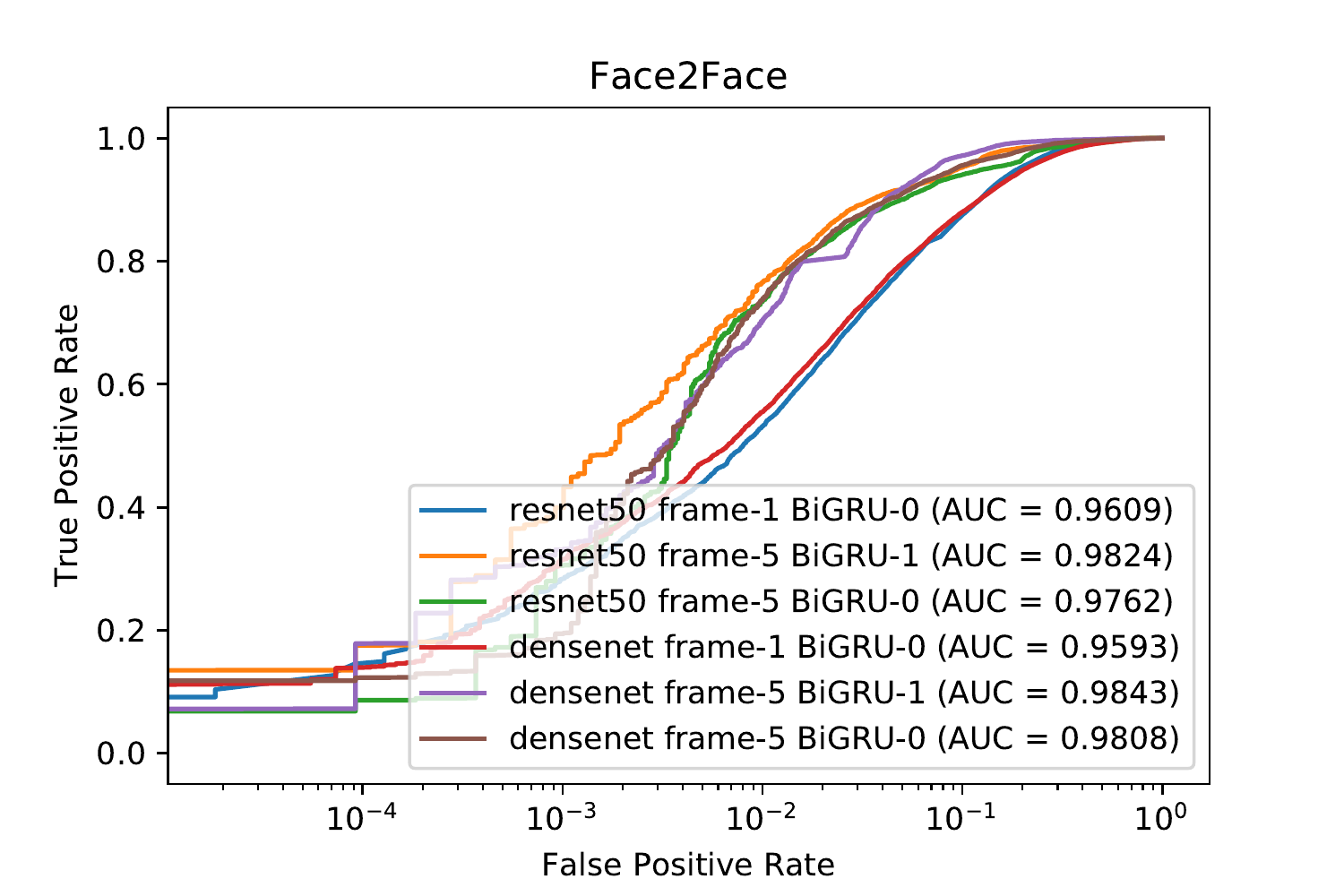}}\\
    \subfloat[]{\includegraphics[width=0.45\linewidth]{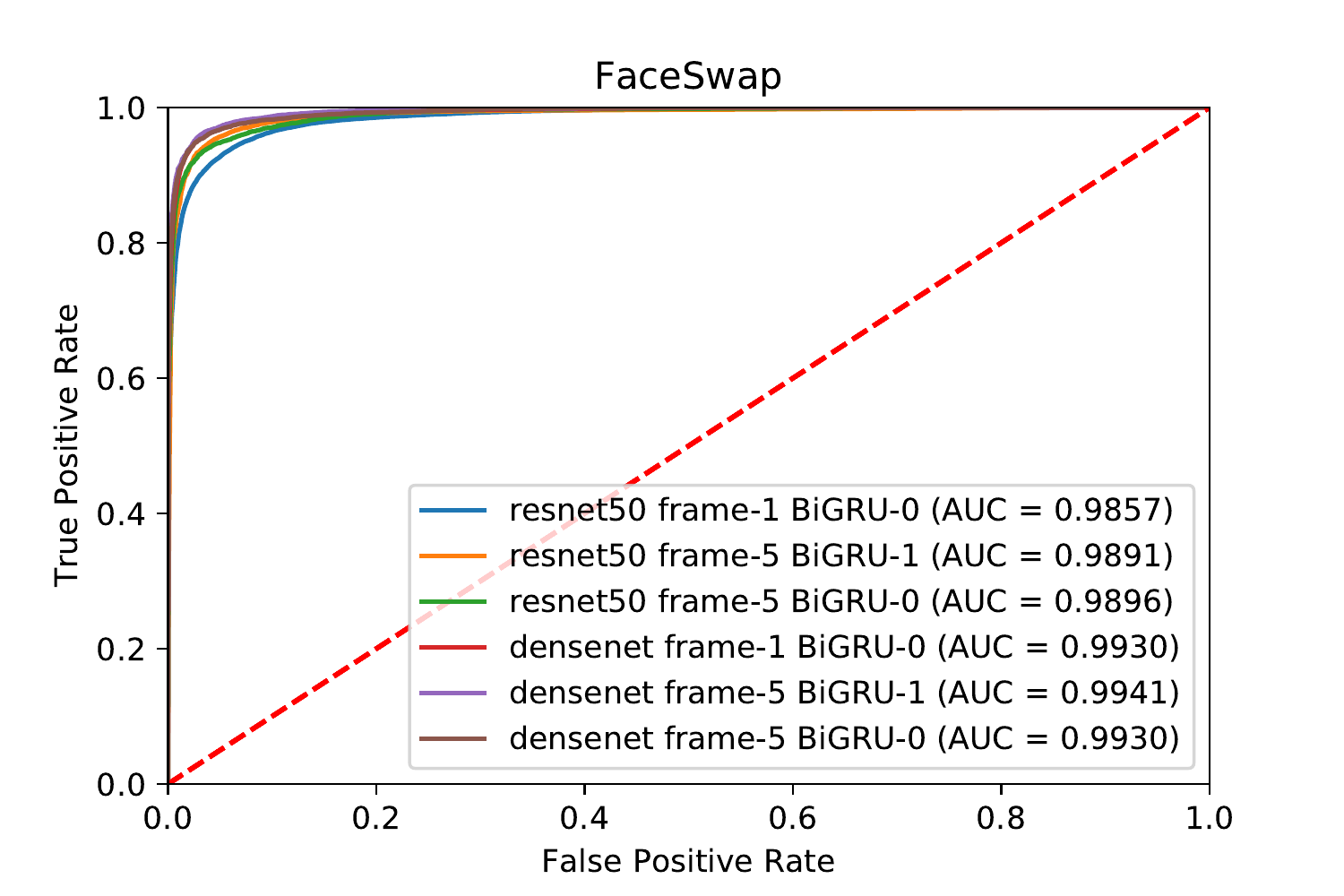} }
    \subfloat[]{\includegraphics[width=0.45\linewidth]{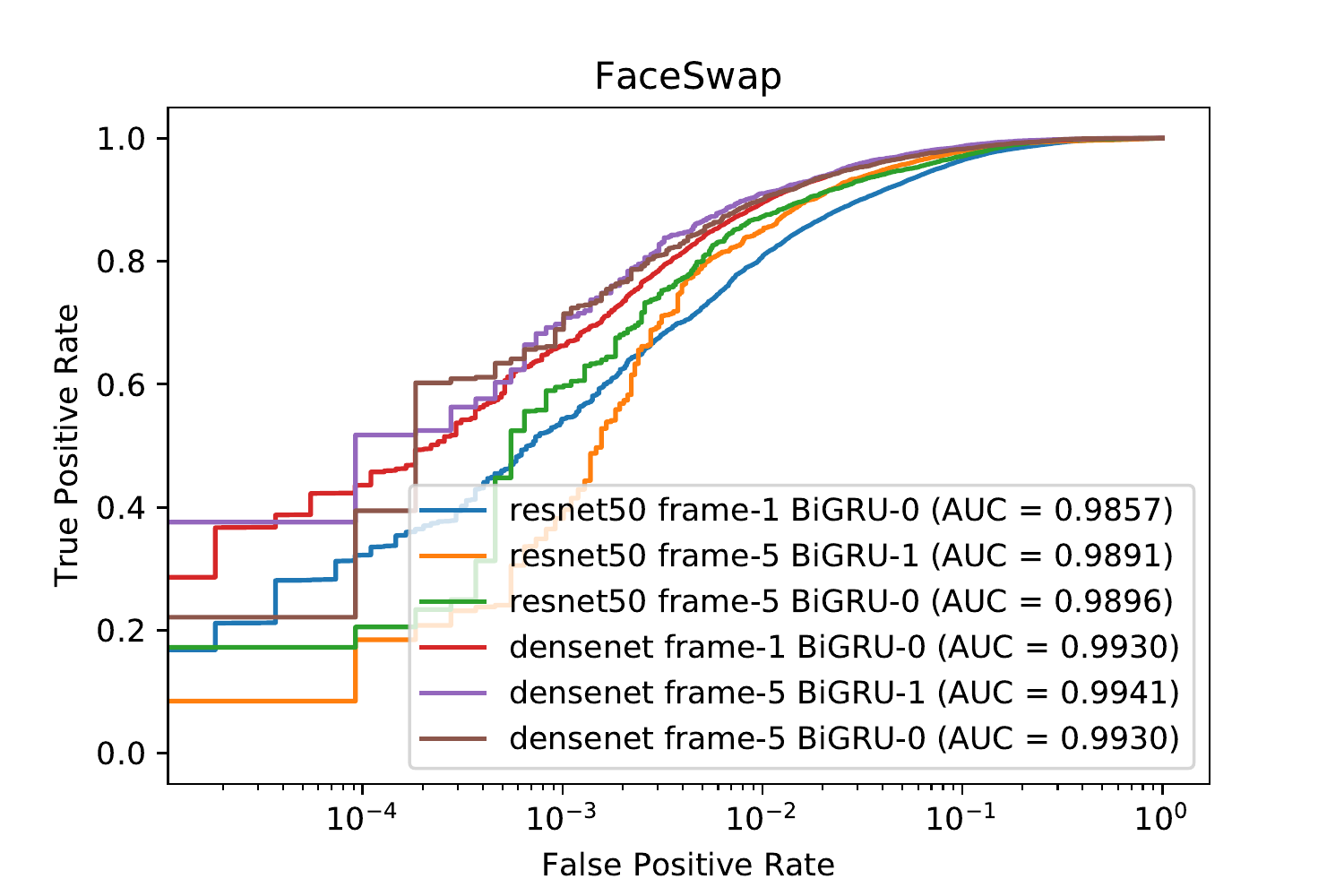}}
    \caption{ROC plots for all manipulation types. Each row corresponds to a different manipulation type. The left column is a linear plot, while the right column has linear-log plots to better analyze the false alarm region. The legend reports ablation studies performed in our experiments: in particular, each experiment describes the backbone network used, how many frames are employed, if bidirectional GRU cells are used or not; finally each item offers the AUC.}
    \label{tab:ROC}
\end{figure*}

\begin{figure*}[!t]
    \centering
    \subfloat[]{\includegraphics[width=0.33\linewidth]{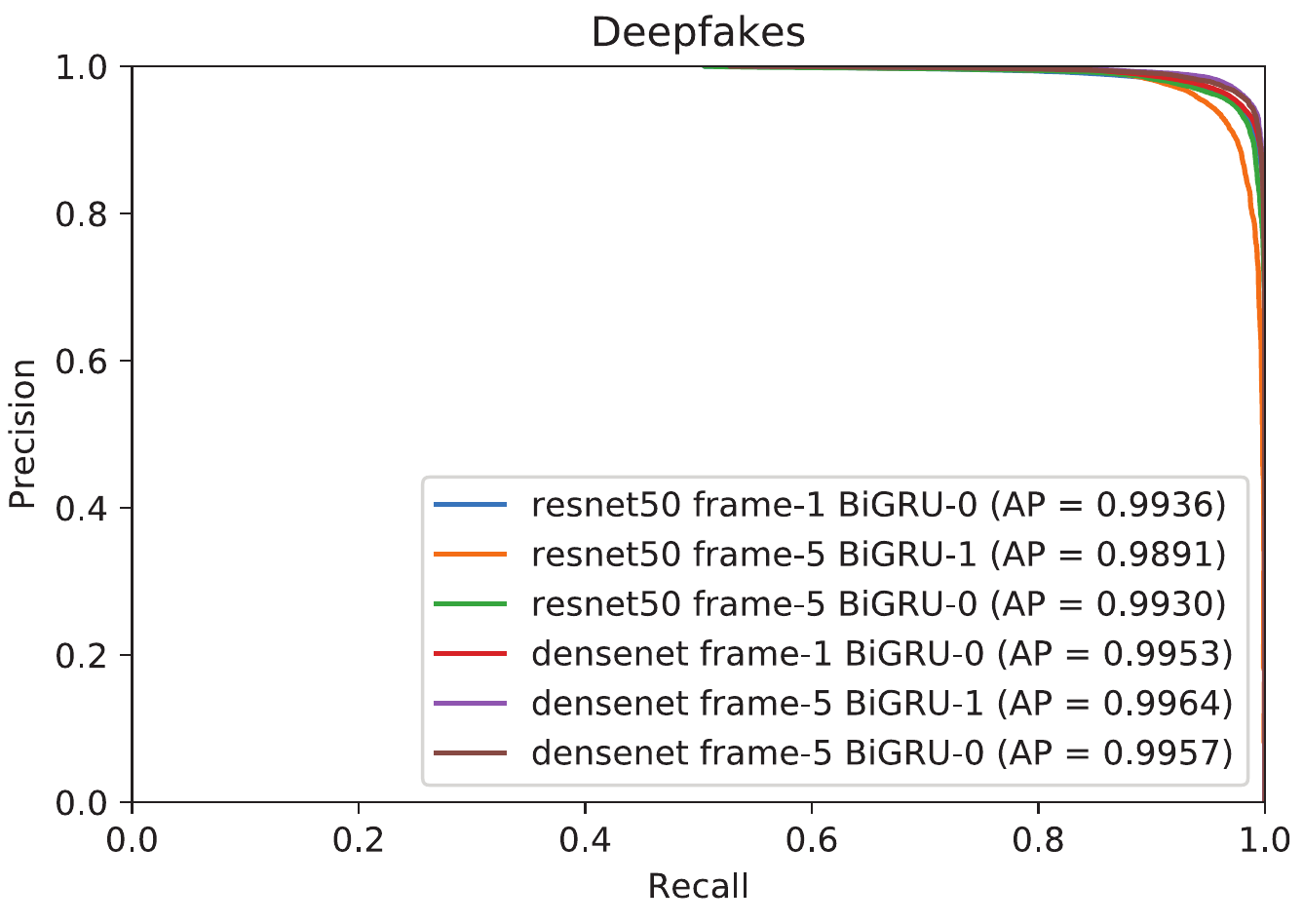} } 
    \subfloat[]{\includegraphics[width=0.33\linewidth]{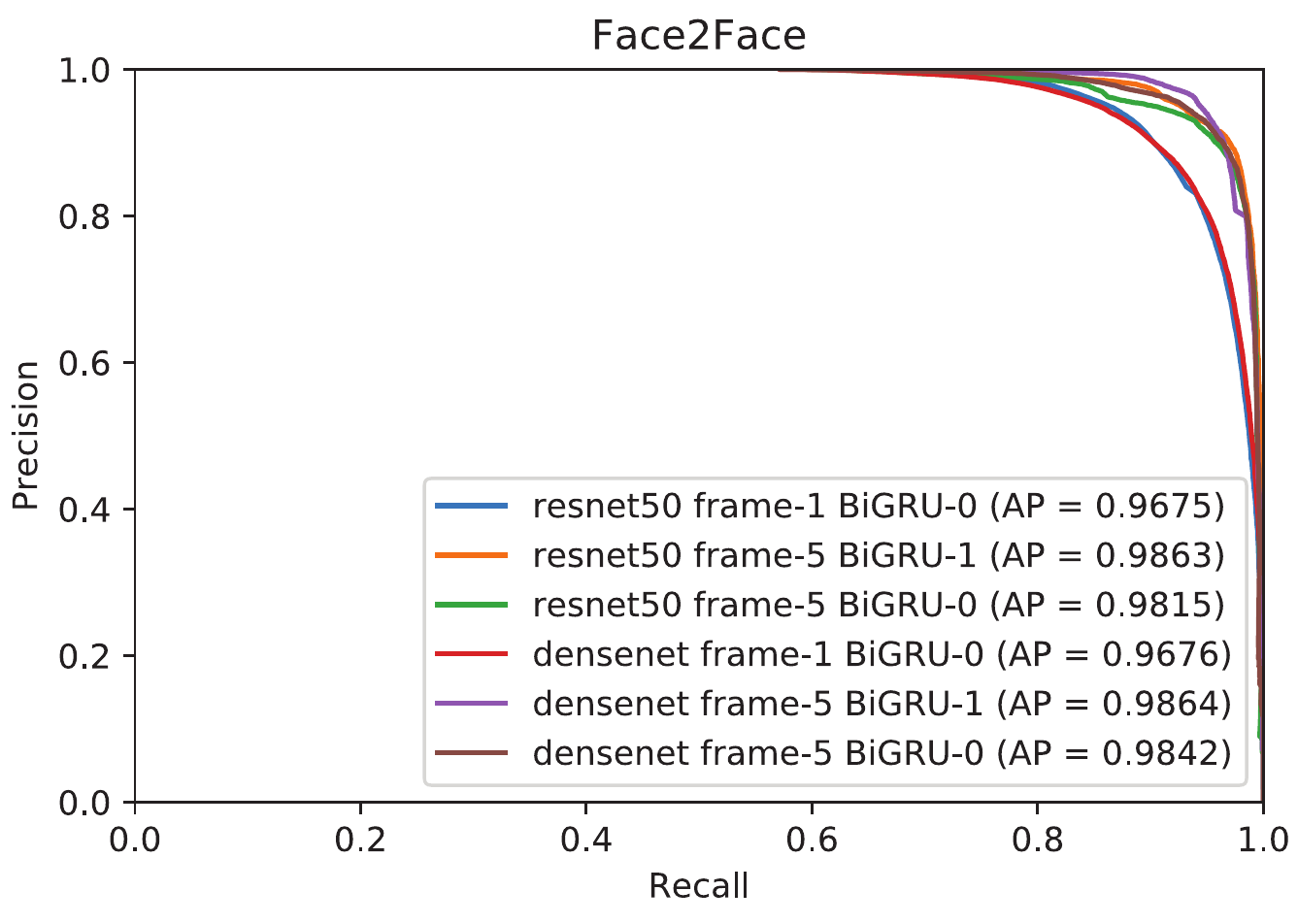}}
    \subfloat[]{\includegraphics[width=0.33\linewidth]{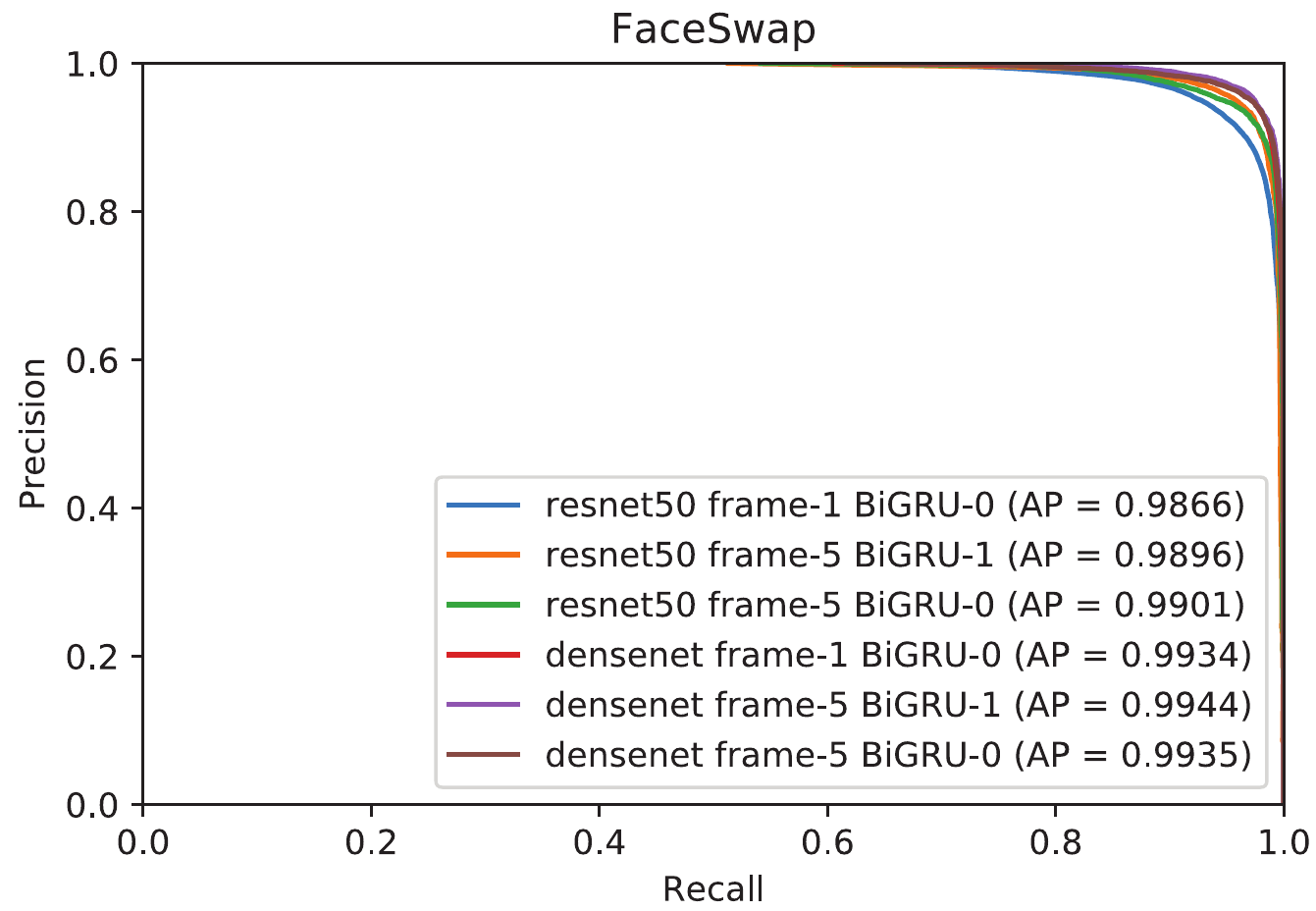} } 
    \caption{Precision Recall curves for all manipulation types. Each column corresponds to a different manipulation type. The legend is similar to \cref{tab:ROC}. Each item also offers the average precision (AP) score.}
    \label{tab:PR}
\end{figure*}

\subsection{Video-based Face Manipulation Detection}\label{sec:video-man}
%\subsection{Bidirection Recurrent Model for Face Manipulation Detection in Videos}\label{sec:video-man}
For manipulation detection, we use a recurrent-convolutional network similar to \cite{donahue_long-term_2015,guera2018deepfake}, where the input is a sequence of frames from the query video. The intuition behind this model is to exploit temporal discrepancies across frames. Temporal discrepancies are expected to occur in images, since manipulations are performed on a frame-by-frame basis. As such, low level artifacts caused by manipulations on faces are expected to further manifest themselves as temporal artifacts with inconsistent features across frames. There are two differences from \cite{donahue_long-term_2015} in our implementation: (1) instead of using CaffeNet \cite{jia_caffe:_2014}, we explore more suitable CNN architectures for the problem, and (2) instead of averaging recurrent features across all time-steps, we extract the final output of the recurrent network. Our work is also different from \cite{guera2018deepfake}, in that we train our model end-to-end, whereas they use pre-trained CNNs. \cref{Fig:Model} shows the model diagram.

\minisection{Backbone encoding network} In our experiments, we explore ResNet \cite{he_deep_2016} and DenseNet \cite{huang_densely_2017} for the CNN component of the model. There are two reasons for exploring these CNNs. FaceForensics++ \cite{rossler_faceforensics++:_2019} is a low resource dataset with a 1,000 videos and to avoid overfitting, authors had to use pre-trained XceptionNet \cite{chollet_xception:_2017} with fixed feature extraction layers. For end-to-end trainability, we chose ResNet \cite{he_deep_2016} which was shown to be easily trainable by the authors. Additionally, manipulation artifacts exhibit low level features (such as discontinuous jawlines, blurred eyes etc.) which do not require high level face semantic features. DenseNet is also a suitable CNN architecture, because it extracts features at different levels of hierarchy \cite{huang_densely_2017}.

Regardless of the architecture used, the backbone network is firstly trained on the FF++ training split minimizing cross-entropy loss for binary classification to develop features to discern real faces from synthetics. The backbone is then extended with RNN and finally trained end-to-end under multiple strategies.

\minisection{RNN training strategies}  In our approach we experiment with the recurrent models placed at different locations of the backbone network: it connects together the backbone network to act as a feature learner that passes features to the RNN aggregating inputs over time. The final system is ultimately trained end-to-end.
We experiment with two strategies: the first simply uses a single recurrent network on top of the final features from the backbone network; alternatively we attempt to learn multiple recurrent networks at different level of the hierarchy of the backbone net: in \cite{afchar_mesonet:_2018}, the authors propose a model that exploits features at a mesoscopic level. Their intuition is that purely microscopic and macroscopic features are not well suited for the face manipulation detection task. We incorporate this idea into our framework by extracting features at multiple feature levels from CNNs for manipulation detection. These features are processed in individual recurrent networks. We expect this new multi-recurrent-convolutional model to utilize micro, meso and macroscopic features for manipulation detection.

\section{Experiments}\label{sec:expts}

%%%%%%%%%%% Subfloat version but it seems it takes more space so reverting back to table.

%%%%%%%%%%%%%%%%%%%%%%%%%
% \begin{table*}[htb]
%     \centering
%     \begin{tabular}{cc}
%         \includegraphics[width=0.9\columnwidth]{images/Deepfakes_full.png} & \includegraphics[width=0.9\columnwidth]{images/Deepfakes_closeup.png} \\
%         \includegraphics[width=0.9\columnwidth]{images/Face2Face_full.png} & \includegraphics[width=0.9\columnwidth]{images/Face2Face_closeup.png} \\
%         \includegraphics[width=0.9\columnwidth]{images/FaceSwap_full.png} & \includegraphics[width=0.9\columnwidth]{images/FaceSwap_closeup.png} \\
%     \end{tabular}
%     \caption{ROC plots for all manipulation types. Each row corresponds to a different manipulation type. The left column is a linear plot, while the right column has semi-log plots to better analyze the false alarm region.}
%     \label{tab:ROC}
% \end{table*}

Our evaluation metric is accuracy for a fair comparison with baselines in \cite{rossler_faceforensics++:_2019}. Additionally, we report area under the receiver operating curve (AUC) scores. All numbers are reported on FaceForensics++. For training, we use Adam optimizer with 1e-4 learning rate. We use GRU cells \cite{cho_learning_2014} for our recurrent network. Additionally, all results are on the heavily compressed version of dataset from \cite{rossler_faceforensics++:_2019}. We do not evaluate on high and low quality videos since the baseline performance for those is already above 98\% in \cite{rossler_faceforensics++:_2019}.

\cref{tab:Table1} shows our results on \textit{Deepfakes}, \textit{Face2Face} and \textit{FaceSwap} manipulation. These results show the impact of variations in architecture. We perform experiments to validate if face alignment improves performance. We use the landmark based face alignment discussed in Section \ref{sec:preproc}. We also check if the temporal aspect of videos helps improve performance. We validate this with two variations: comparing five frame vs single frame input and uni- vs bi-directional recurrent network for five frame input experiments. We use a single recurrent network here on top of final CNN features. We consistently find (1) DenseNet to outperform ResNet, (2) face alignment to give improvement and (3) a sequence of images to be better than single frame input. We also find bidirectional recurrence to be superior to uni-directional recurrence. \cref{tab:ROC} and \cref{tab:PR} report ROC and precision-recall (PR) curves for the same results. Since the false alarm region is hard to discern, we also show the linear-log plots for the ROC curve.

\cref{tab:Table2} shows our results from the choice of face alignment method and multiple levels of recurrence of our model. We use the best model from \cref{tab:Table1}, namely the five-frame-bidirectional-densenet as our base model. The first variation is to use a spatial transformer network (STN) for learning an affine alignment of faces instead of a landmark based alignments as discussed in Section \ref{sec:preproc}. For our experiments, we use a simple CNN with 2 convolution and max-pooling layers each followed by a feedforward network as the localization net and bilinear interpolation for the sampler. The second variation is the multi-recurrent model which involves extracting recurrent features at multiple levels of CNNs as explained in Section \ref{sec:video-man}. Specifically, since the DenseNet used in our experiments has four blocks for generating feature maps, we build four recurrent networks for these. We find both strategies to be unsuccessful for improving performance. With respect to STN, we notice that training is relatively unstable and this may be due to changes in affine parameters during training which impacts the performance of the following CNN network. This may be a possible reason for performance drop when using STN. A possible reason for the relatively poor performance of the multi-recurrence model is the significant increase in the number of parameters, which leads to overfitting, given that the FaceForensics++ dataset has a limited number of samples (1,000 videos).

%-------------------------------------------------------------------------
\section{Conclusion}\label{sec:conclusions}
Misinformation in online content is increasing and there is an exigent need for detecting such content. Face manipulation in videos is one aspect of the larger problem. In this work we showed that a combination of recurrent-convolutional model and face alignment approach improves upon the state-of-the-art. We also explored different strategies for both alignment and combining CNN features through recurrence. We found a landmark based face alignment with bidirectional-recurrent-denset to perform the best for face manipulation detection in videos.

\section*{Acknowledgments}
This work is based on research sponsored by the Defense Advanced Research Projects Agency under agreement number FA8750-16-2-0204. The U.S. Government is authorized to reproduce and distribute reprints for governmental purposes notwithstanding any copyright notation thereon. The views and conclusions contained herein are those of the authors and should not be interpreted as necessarily representing the official policies or endorsements, either expressed or implied, of the Defense Advanced Research Projects Agency or the U.S. Government.

\balance

{\small
\bibliographystyle{ieee}
\bibliography{FaceForensics}
}

\end{document}